\documentclass[11pt]{article}

\usepackage[preprint]{acl}

\usepackage{times}
\usepackage{latexsym}

\usepackage[T1]{fontenc}

\usepackage{multicol}

\usepackage{xspace}
\usepackage{booktabs}
\usepackage{pifont}

\newcommand{\system}{\textsc{ViviDoc}\xspace}
\usepackage{multirow}
\usepackage{makecell}

\usepackage[utf8]{inputenc}

\usepackage{microtype}
\usepackage[ruled,lined,linesnumbered]{algorithm2e}
\usepackage{amsmath}
\usepackage{amssymb}
\usepackage{inconsolata}

\usepackage{graphicx}
\usepackage{booktabs}  
\usepackage{tabularx}  
\usepackage{ragged2e}  

\usepackage{multirow}
\usepackage{xcolor}   
\usepackage{colortbl} 
\usepackage{arydshln} 
\usepackage{fontawesome5} 
\usepackage{graphicx}

\definecolor{bestblue}{RGB}{220, 235, 250}  
\definecolor{secondgreen}{RGB}{235, 245, 230} 

\usepackage{enumitem}

\title{\sc{Demonstrating ViviDoc: Generating Interactive Documents through Human-Agent Collaboration}}

\author{
\normalfont
Yinghao Tang\textsuperscript{1},
Yupeng Xie\textsuperscript{2},
Yingchaojie Feng\textsuperscript{3},
Tingfeng Lan\textsuperscript{4},
Wei Chen\textsuperscript{1}
\\
\normalsize
\textsuperscript{1}\textit{State Key Lab of CAD\&CG, Zhejiang University} ~~~
\textsuperscript{2}\textit{HKUST(GZ)} \\
\textsuperscript{3}\textit{National University of Singapore} ~~~
\textsuperscript{4}\textit{University of Virginia}
\normalsize
\\
\normalsize
}

\begin{document}
\maketitle
\begin{abstract}
Interactive articles help readers engage with complex ideas through exploration, yet creating them remains costly, requiring both domain expertise and web development skills. Recent LLM-based agents can automate content creation, but naively applying them yields uncontrollable and unverifiable outputs. We present \system, a human-agent collaborative system that generates interactive educational documents from a single topic input. \system introduces a multi-agent pipeline (Planner, Executor, Evaluator) and the Document Specification (DocSpec), a human-readable intermediate representation that decomposes each interactive visualization into State, Render, Transition, and Constraint components. The DocSpec enables educators to review and refine generation plans before code is produced, bridging the gap between pedagogical intent and executable output. Expert evaluation and a user study show that \system substantially outperforms naive agentic generation and provides an intuitive editing experience. Our project homepage is available at \url{https://vividoc-homepage.vercel.app/}.
\end{abstract}    
\section{Introduction}

\begin{figure}[t]
    \centering
    \includegraphics[width=\columnwidth]{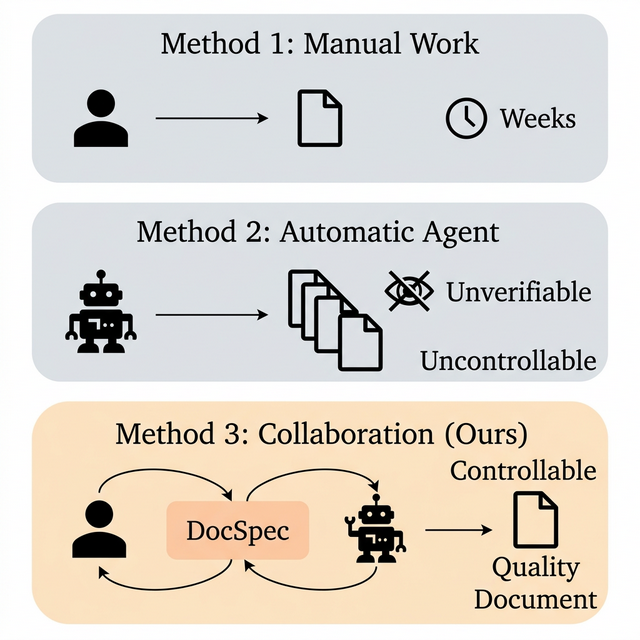}
    \caption{ Three approaches to creating interactive documents: manual authoring, fully automatic generation, and human-agent collaboration through DocSpec.}
    \label{fig:teaser}
\end{figure}

Interactive articles are an emerging communication medium that leverages the dynamic capabilities of the web to help readers engage with complex ideas~\cite{victor2011explorable}. Through interactive elements such as sliders, dropdowns, and direct manipulation controls, readers can actively explore concepts, observe cause-and-effect relationships, and build intuition through experimentation. This form of communication has been adopted across diverse domains: explorable explanations in education~\cite{chi2014icap}, data-driven storytelling in journalism, and interactive tutorials in scientific publishing~\cite{hohman2020communicating}. Studies show that interactive articles improve reader engagement and learning compared to static alternatives~\cite{hohman2020communicating}.

Despite their promise, creating interactive articles remains prohibitively costly. Producing a single piece, such as those on Explorable Explanations or Distill.pub \cite{hohman2020communicating}, requires expertise in both domain knowledge and web development, often demanding days or weeks of effort per article. This bottleneck severely limits the availability of high-quality interactive content across domains.

Recent advances in LLM-based agents have shown promise in automating complex content creation tasks such as video generation \cite{chen2025code2video}, slide design \cite{zheng2025pptagent, liang2025slidegen}, and infographic production \cite{tang2026igenbench}. These systems combine the text and code generation capabilities of large language models (LLMs) with agentic planning and tool use. However, applying such approaches to interactive educational documents remains largely unexplored, and doing so naively faces three key limitations. First, the generation process is \textbf{largely uncontrollable}: there is a fundamental gap between pedagogical intent (what the educator wants the learner to experience) and the executable code that realizes that experience, and the agent resolves this gap based on its own implicit preferences. Second, humans cannot effectively participate in the process: the agent operates as a \textbf{black box}, leaving educators no meaningful way to review or adjust the intermediate decisions that shape the final output. Third, datasets for this task are \textbf{scarce}, making it difficult to systematically evaluate approaches.

To address these limitations, we present \system, a human-agent collaborative system for generating interactive educational documents. \system is built on two key ideas. First, we design a multi-agent pipeline where a Planner agent decomposes a topic into a structured plan, an Executor agent synthesizes text and code from the plan, and an Evaluator agent checks the output for correctness. Second, we introduce the Document Specification (DocSpec), a structured intermediate representation that organizes a document into knowledge units, each containing a text description for content generation and an Interaction Specification for visualization generation. The Interaction Specification, developed from interactive visualization theory~\cite{munzner2014visualization}, decomposes each visualization into four components: State, Render, Transition, and Constraint (SRTC). We extend this theoretical framework to serve as a controllable intermediate layer for LLM-based generation. Because the DocSpec is structured and human-readable, educators can review, edit, and refine it before code is produced, enabling meaningful human control over the generation process. The DocSpec also acts as a contract between agents, reducing ambiguity at the most error-prone step: translating intent into code.

We summarize our contributions as follows:
\begin{itemize}[noitemsep,leftmargin=10pt]
    \item We design a multi-agent system that generates interactive educational documents from a single topic input through structured planning, code synthesis, and automated evaluation.
    
    \item We propose the Document Specification (DocSpec), a structured intermediate representation consisting of text descriptions and Interaction Specifications (SRTC), that makes the generation process controllable and allows human intervention before code is produced.
    
   \item We collect a dataset of 101 real-world interactive documents from over 60 websites across 11 domains. Expert blind evaluation on 10 randomly sampled topics shows that \system substantially outperforms naive agentic generation in content richness, interaction quality, and visual quality, and a user study confirms that the DocSpec editing interface is easy to learn and effective for guiding the generation process.

\end{itemize}

\section{Related Work}
\label{sec:related} 

\subsection{Interactive Document}
Interactive documents, often conceptualized as explorable explanations~\cite{victor2011explorable}, are a transformative medium that leverages dynamic web capabilities to foster deep engagement with complex ideas. The theoretical foundation for such active reading environments traces back to early human-computer interaction paradigms, including Engelbart's framework for augmenting human intellect~\cite{engelbart2023augmenting} and foundational hypertext structures~\cite{nelson1965complex}, which were operationalized in early systems like PLATO~\cite{bitzer2007plato}. Today, empirical studies confirm that interactive articles significantly improve learner engagement and comprehension compared to static alternatives~\cite{chi2014icap, hohman2020communicating}. The medium has been successfully adopted across diverse fields, from immersive data journalism (e.g., The New York Times' Snow Fall~\cite{branch2012snow} and Bloomberg's climate visualizations~\cite{roston2015whats}) to scholarly communications exploring machine learning fairness~\cite{wattenberg2016attacking}.

Despite their efficacy, creating these artifacts remains prohibitively costly, demanding a rare intersection of deep domain knowledge and web development expertise. This bottleneck severely limits the availability of high-quality interactive content. To overcome these authoring barriers, we introduce \system, a human-agent collaborative system designed to structurally automate and control the generation of interactive educational documents.

\subsection{LLM Agents for Content Creation}

Recent advancements increasingly rely on LLMs and multi-agent systems to automate complex content creation by decomposing tasks across specialized agents \cite{lin2025creativity}. These frameworks have achieved success across various domains, including data visualization systems like LIDA \cite{dibia2023lida}, Infogen \cite{ghosh2025infogen}, HAIChart\cite{xie2024haichart}, and PlotGen \cite{goswami2025plotgen} that synthesize complex infographics from unstructured text \cite{tang2026igenbench}. Similarly, presentation agents like PPTAgent \cite{zheng2025pptagent}, and SlideGen \cite{liang2025slidegen} use structural schemas for coherent slide decks, while systems like LAVES \cite{yan2026beyond} extend multi-agent orchestration to generate synchronized educational videos.

However, applying such agentic approaches to interactive educational documents remains largely unexplored. Naive implementation faces critical limitations regarding uncontrollable generation lacking pedagogical alignment, the black-box nature of agents that preclude meaningful human intervention, and a scarcity of specialized datasets for systematic evaluation. Addressing these fundamental bottlenecks, \system establishes a transparent, human-in-the-loop generation paradigm by structuring the collaborative process around an interpretable intermediate representation.

\section{System Overview}
\label{sec:system}

\begin{figure*}[t]
\centering
\includegraphics[width=0.9\textwidth]{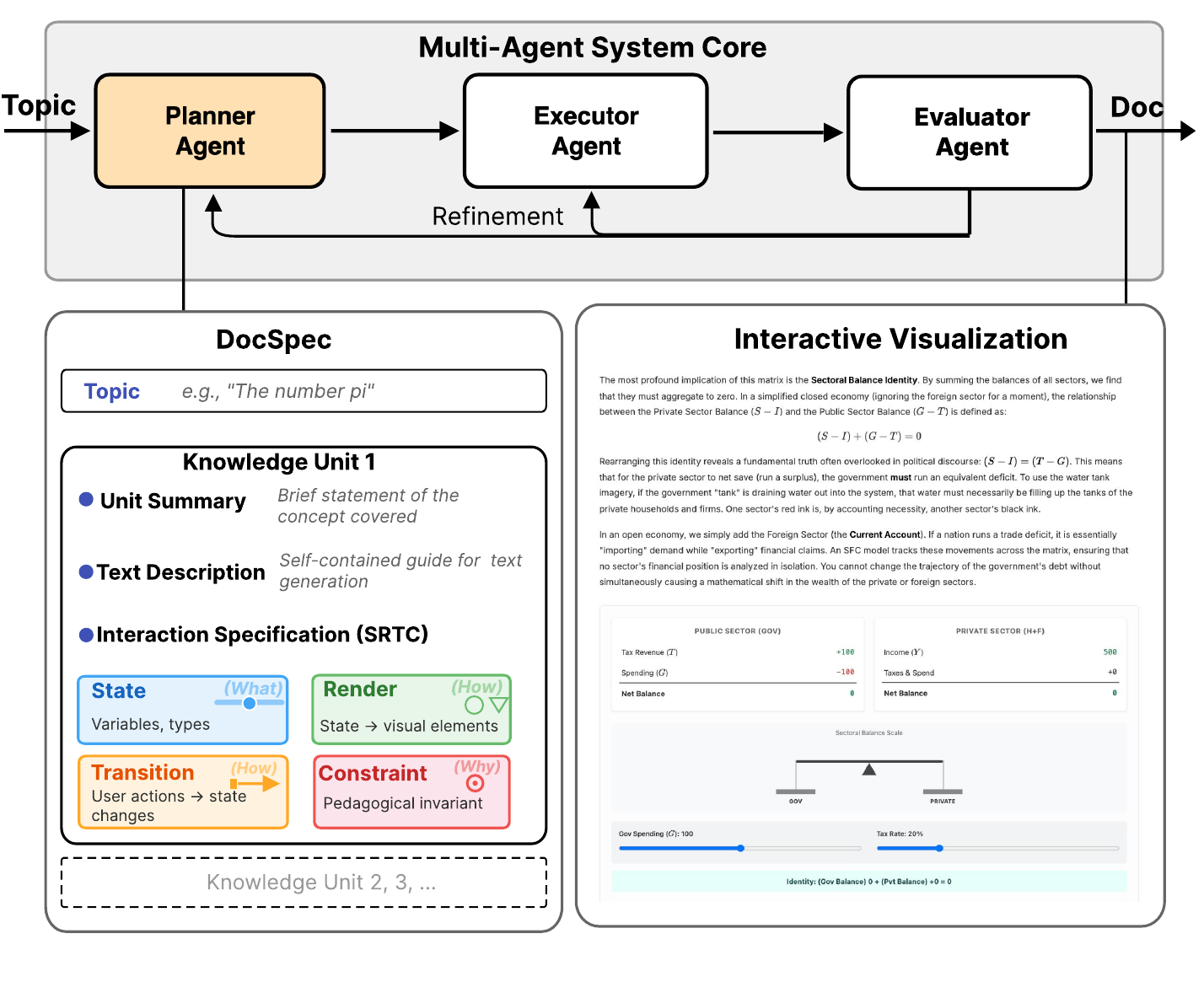}
\caption{The \system pipeline. The Planner generates a Document Specification (DocSpec) from a topic. The user can review and edit the DocSpec before the Executor generates the final document. The Evaluator checks the output and provides feedback.}
\label{fig:pipeline}
\end{figure*}

Figure~\ref{fig:pipeline} illustrates the \system pipeline. Given a topic as input, \system generates a complete interactive educational document through three agents coordinated by a structured intermediate representation, the Document Specification (DocSpec). A key design principle is that the DocSpec is exposed to the user between planning and execution, enabling human review and editing before any code is generated.

\subsection{Pipeline}

\paragraph{Planner.} The Planner agent takes a topic (e.g., ``What is the rank of a matrix?'') and produces a DocSpec. It decomposes the topic into a sequence of knowledge units, each containing a text description that guides content generation and an Interaction Specification (SRTC) that defines the interactive visualization. The Planner uses an LLM with structured output to ensure the DocSpec conforms to a predefined schema. We describe the DocSpec structure in detail in Section~\ref{sec:spec}.

\paragraph{Executor.} The Executor agent takes the (possibly edited) DocSpec and generates the final HTML document. It processes each knowledge unit in two stages. In Stage 1, it generates the text content as an HTML fragment, guided by the text description and the context of previously generated sections to maintain stylistic consistency. In Stage 2, it generates the interactive visualization as HTML, CSS, and JavaScript, guided by the SRTC Interaction Specification. The Executor includes a retry mechanism: if a generated fragment fails validation, it regenerates with up to $k$ attempts.

\paragraph{Evaluator.} The Evaluator agent checks the generated document for quality. It assesses overall text coherence and logical flow using an LLM, and verifies that all knowledge units have been successfully generated and pass HTML validation. If issues are found, the Evaluator provides feedback that can trigger re-execution of specific components.

\paragraph{Human Review.} Human review is not limited to a single stage but spans the entire pipeline. After the Planner produces the DocSpec, the user can reorder knowledge units, modify text descriptions, adjust interaction parameters (e.g., variable ranges, control types), or add and remove units entirely. Because the DocSpec is structured rather than free-form, edits are targeted and predictable: changing a slider range in the Interaction Specification directly affects the generated visualization without requiring the user to write code. After the Executor produces the final document, the user can further review the output and request revisions. In both stages, the user may either edit directly through the interface or converse with an LLM-powered chat assistant to describe desired changes in natural language. This step is optional but provides human control at the points where it is most effective.

The DocSpec serves as a contract between pipeline stages: the Planner expresses pedagogical intent in a structured form, the Executor implements that intent as code, and the Evaluator verifies the result. This decomposition isolates the most error-prone step, the translation from intent to code, and constrains it with a structured specification rather than ambiguous natural language. Placing human review at both the planning and output stages ensures that editorial decisions can be made where they have the highest leverage, before code is generated, and where they are most visible, after the final document is produced.

\begin{figure}[t!]
\centering
\includegraphics[width=\columnwidth]{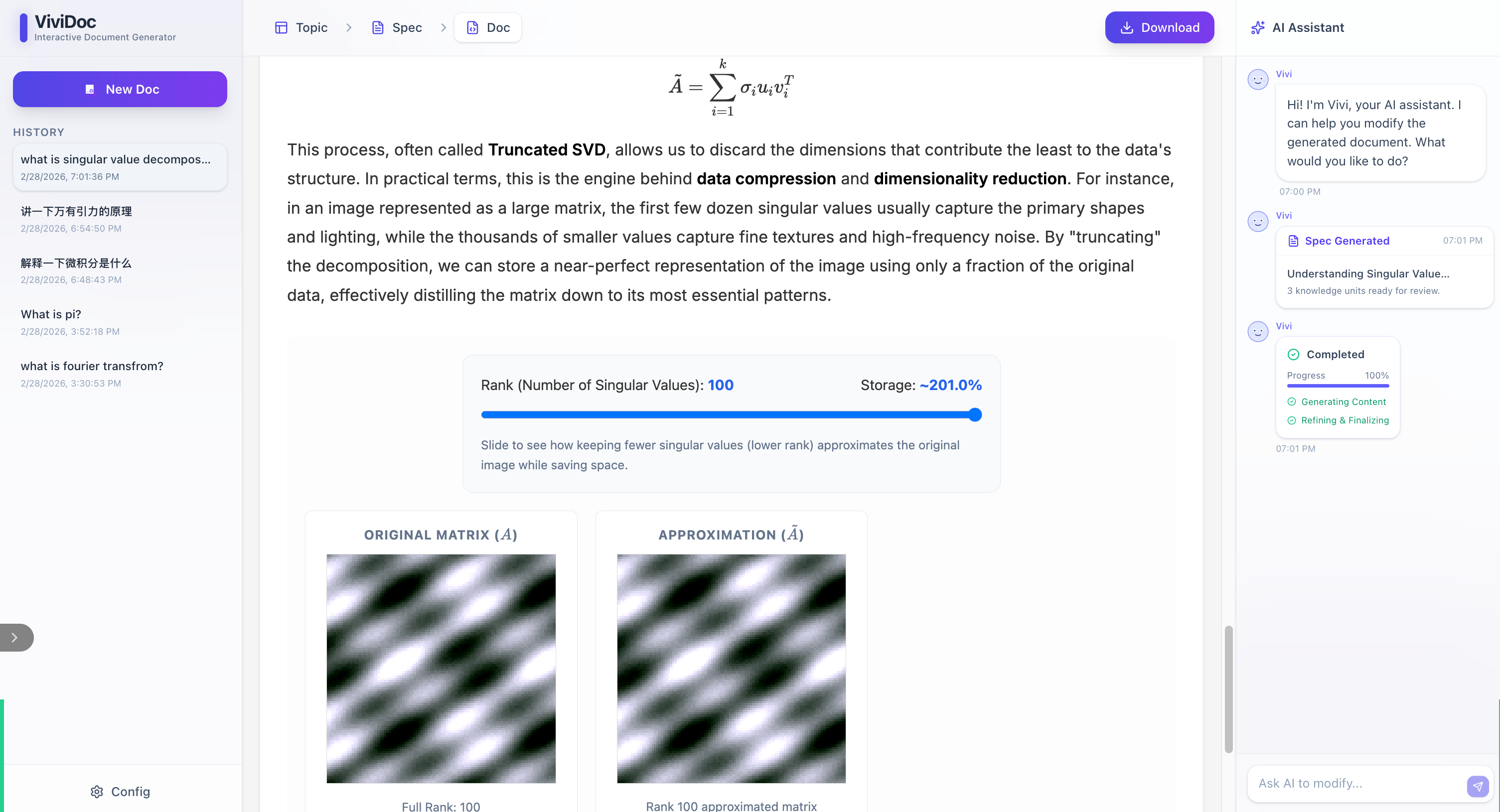}
\caption{The \system user interface. Users follow a Topic $\rightarrow$ Spec $\rightarrow$ Doc workflow to generate interactive educational documents.}
\label{fig:demo}
\end{figure}

\section{Document Specification}
\label{sec:spec}

The Document Specification (DocSpec) is the structured intermediate representation at the core of \system. It bridges the gap between a high-level topic and the generated interactive document by decomposing the content into a sequence of knowledge units, each with explicit instructions for both text and interaction generation.

\subsection{Structure}

A DocSpec consists of a topic and an ordered list of knowledge units. Each knowledge unit contains three components:

\begin{itemize}[noitemsep,leftmargin=10pt]
    \item A \textbf{unit summary} that briefly states the concept covered.
    \item A \textbf{text description} that provides a self-contained guide for generating the explanatory text of the section. It specifies what the reader should understand after reading, without prescribing exact wording.
    \item An \textbf{Interaction Specification} that defines the interactive visualization using the SRTC decomposition described below.
\end{itemize}

The text description and Interaction Specification serve different stages of the Executor: the former guides Stage 1 (text generation) and the latter guides Stage 2 (visualization generation). This separation allows each stage to operate independently with minimal ambiguity. The Interaction Specification supports common interaction patterns such as parameter exploration (sliders), selection (dropdowns), direct manipulation (drag), and mode switching (toggles), while static visualizations are handled as a degenerate case where no transitions are defined.

\subsection{Interaction Specification: SRTC}

Munzner's What-Why-How framework~\cite{munzner2014visualization} provides a foundational decomposition for visualization design: What data is being visualized, How it is visually encoded and interacted with, and Why the user engages with it. We adapt this framework to the setting of LLM-based generation by introducing the SRTC Interaction Specification, which decomposes each interactive visualization into four components:

\begin{itemize}[noitemsep,leftmargin=10pt]
    \item \textbf{State (S)}: The variables underlying the visualization, including their types, domains, default values, and derivation rules. Variables are either user-controllable (e.g., a slider with a specified range) or derived from other variables (e.g., a formula). This corresponds to Munzner's What.
    \item \textbf{Render (R)}: A description of how the state maps to visual elements on screen, such as geometric shapes, labels, or charts. This corresponds to the visual encoding aspect of How.
    \item \textbf{Transition (T)}: A description of how user actions modify the state, specifying the cause-and-effect relationship between input events and state changes. This corresponds to the interaction idiom aspect of How.
    \item \textbf{Constraint (C)}: The pedagogical invariant that the visualization is designed to demonstrate. This is the key insight the learner should discover through interaction. This corresponds to Munzner's Why, adapted from ``what task is the user performing'' to ``what should the learner observe.''
\end{itemize}

This structured format benefits all three agents: it constrains the Planner to produce well-formed specifications, provides the Executor with an unambiguous contract for code synthesis, and gives the Evaluator an objective criterion (the Constraint C) for verifying correctness.

\paragraph{Example.} Table~\ref{tab:spec-example} contrasts a natural language interaction description with its corresponding SRTC specification for a visualization about $\pi$. The natural language version leaves implicit what variables exist, how they relate, what appears on screen, and what invariant the learner should notice. The SRTC version makes each of these explicit. The State defines a user-controllable variable $r$ (a slider over $[0.5, 5]$) and three derived variables: circumference $C = 2\pi r$, diameter $D = 2r$, and their ratio $C/D$. The Render specifies that the visualization displays a circle whose size reflects $r$, along with labels for $C$, $D$, and the ratio. The Transition links the slider interaction to the state: dragging the slider changes $r$, and all derived variables update automatically. Finally, the Constraint encodes the pedagogical invariant that the learner should discover: $C/D \approx 3.14159$ regardless of $r$.

\begin{table}[t]
\small
\centering
\begin{tabularx}{\columnwidth}{@{}lX@{}}
\toprule
\multicolumn{2}{@{}l}{\textbf{Natural language description}} \\
\midrule
\multicolumn{2}{@{}p{\columnwidth}@{}}{``The reader can adjust the radius of a circle using a slider, and as the radius changes, the visualization updates the circle while recalculating and displaying the circumference, diameter, and ratio, allowing the reader to see that the ratio stays roughly the same.''} \\
\midrule
\multicolumn{2}{@{}l}{\textbf{SRTC Interaction Specification}} \\
\midrule
\textbf{S} & \texttt{r}: slider $[0.5, 5]$, default 1; \texttt{C}: derived $2\pi r$; \texttt{D}: derived $2r$; \texttt{ratio}: derived $C/D$ \\
\textbf{R} & A circle whose size reflects $r$; labels showing $C$, $D$, and ratio \\
\textbf{T} & Dragging the slider changes $r$; $C$, $D$, ratio update automatically \\
\textbf{C} & ratio $\approx 3.14159$ regardless of $r$ \\
\bottomrule
\end{tabularx}
\caption{Comparison of a natural language interaction description and its SRTC Interaction Specification for a visualization about $\pi$.}
\label{tab:spec-example}
\end{table}

\section{Experiments}
\label{sec:experiment}

\subsection{Output Quality Evaluation}
\label{sec:study1}

We randomly sampled 10 topics from our benchmark dataset and generated interactive educational documents using two methods, producing 20 documents in total.
\system follows the multi-agent pipeline: the Planner generates a DocSpec, the Executor generates the document, and the Evaluator checks the output.
To ensure a fair comparison, we disable the Human Review step so that no human feedback is involved in either method.
Naive Agent uses the same topic as input and prompts the LLM to generate a complete HTML document in a single call, without DocSpec planning.
Both methods use the same underlying model (Gemini 3.0 Flash) to control for differences in model capability.

We recruited three domain experts with backgrounds in interactive educational content design or visualization research to conduct a blind evaluation of the 20 documents.
Experts were not informed of which method produced each document.
They rated each document on three dimensions using a 5-point Likert scale:
\textit{Content Richness} (whether the document covers the topic with sufficient depth and breadth),
\textit{Interaction Quality} (whether the interactive design helps readers explore and understand the core concept), and
\textit{Visual Quality} (the layout and readability of the document).

\begin{figure}[t]
  \centering
  \includegraphics[width=\columnwidth]{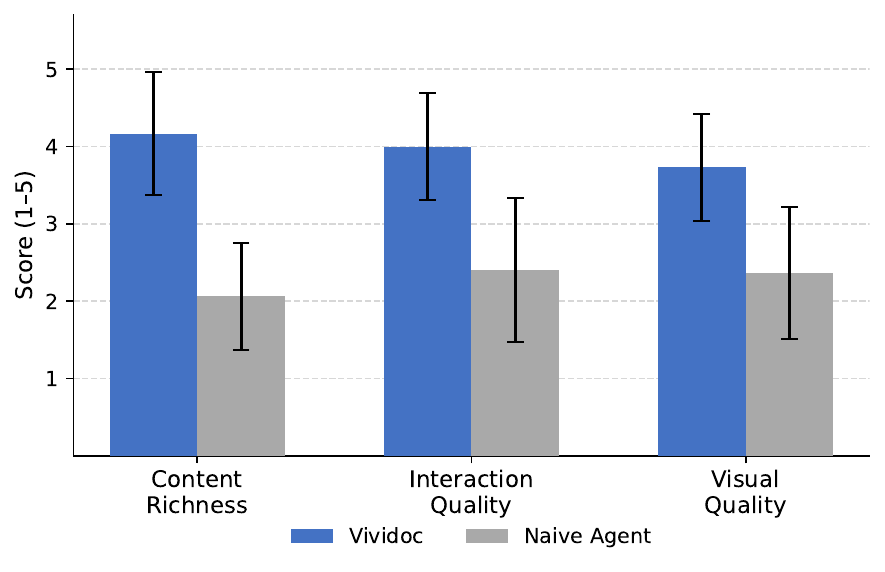}
  \caption{Expert blind evaluation scores for \system and Naive Agent across three dimensions.}
  \label{fig:quality}
\end{figure}

As shown in Figure~\ref{fig:quality}, \system outperforms Naive Agent on all three dimensions.
The gap is largest for Content Richness (4.17 vs.\ 2.07), followed by Interaction Quality (4.00 vs.\ 2.40) and Visual Quality (3.73 vs.\ 2.37).
These results indicate that without structured planning through DocSpec, a single-call LLM struggles to produce documents with well-organized content, coherent interactive design, and clear visual layout.
The most pronounced drop in Content Richness suggests that without structured planning, LLM-generated documents tend to be shallow and narrowly focused, while the drop in Interaction Quality confirms that the SRTC Interaction Specification plays a central role in constraining the generation of meaningful interactions.

\subsection{User Study}
\label{sec:userstudy}

We conducted a user study to evaluate the usability of \system.
The study focused on whether participants could effectively use the DocSpec editing interface to guide the generation process toward their intended output.

\paragraph{Participants.}
We recruited three participants with backgrounds in visualization or educational technology research.
All participants had prior experience in interactive content creation or educational tool design.

\paragraph{Procedure.}
Each participant completed three document generation tasks covering topics in mathematics, algorithms, and physics, presented in a random order to reduce ordering effects.
A 10-minute introduction and practice session was provided before the tasks began.
Tasks had no time limit.
Participants were encouraged to think aloud throughout the session, and a 20-minute semi-structured interview was conducted afterward to collect qualitative feedback.

\begin{figure}[t]
  \centering
  \includegraphics[width=\columnwidth]{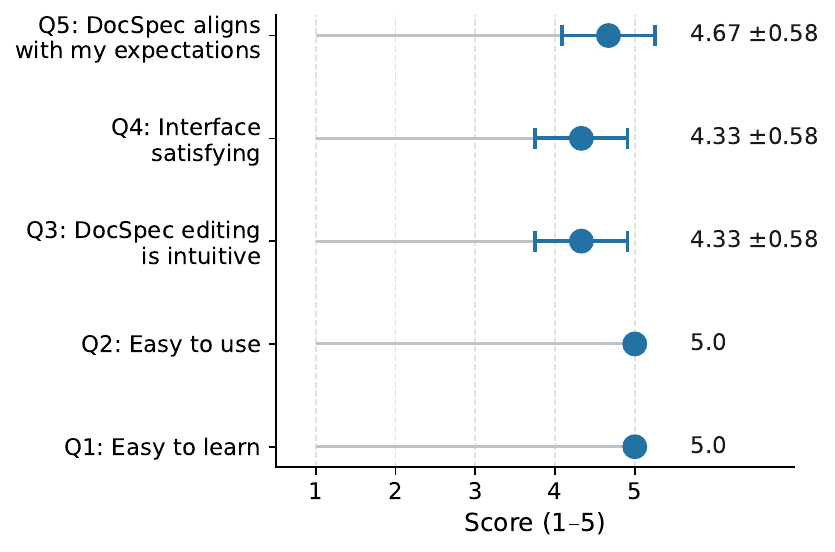}
  \caption{Participant ratings of \system on five usability dimensions. Values show mean and standard deviation.}
  \label{fig:likert}
\end{figure}

\paragraph{Usability Ratings.}
After completing the tasks, participants rated the system on five items using a 5-point Likert scale.
As shown in Figure~\ref{fig:likert}, all items received positive ratings: easy to learn (5.0), easy to use (5.0), DocSpec editing is intuitive (4.33 $\pm$ 0.58), interface satisfying (4.33 $\pm$ 0.58), and DocSpec aligns with my expectations (4.67 $\pm$ 0.58), indicating that the DocSpec interface effectively supports the document generation process.

In the interviews, participants noted that DocSpec was the first time they could ``actually decide how the interaction works without writing any code'' (P1), and that the separation between the text description and the interaction specification made editing more straightforward (P2).
\section{Conclusion}
We presented \system, a human-agent collaborative system for generating interactive educational documents. By introducing the Document Specification (DocSpec) as a structured intermediate representation, \system enables meaningful human control over the generation process while leveraging multi-agent automation. Experiments show that structured planning through DocSpec leads to substantial quality improvements over naive generation, and a user study confirms that the editing interface is intuitive and effective. We hope \system lowers the barrier to creating interactive content and inspires further research on human-AI collaboration in document authoring.

\bibliography{custom}
\newpage
\section{Ethical Considerations}
The dataset of 101 interactive documents was collected and utilized in strict compliance with applicable copyright regulations. For our user study, we obtained explicit informed consent from all participants and rigorously anonymized all interview records to protect personal privacy. Although \system employs LLMs to assist in generation, we mitigate potential risks of AI hallucinations and uncontrollable outputs through a human-in-the-loop paradigm. By allowing users to review and edit the Document Specification (DocSpec) prior to code synthesis, the system ensures that educators retain full control over the pedagogical intent. As this work focuses on collaborative human-agent authoring rather than deploying a fully autonomous system, we do not foresee any significant ethical or societal risks.
\appendix
\end{document}